\documentclass[review]{elsarticle}
\usepackage{hyperref}
\usepackage{float}
\usepackage{verbatim} 
\usepackage{apalike}
\usepackage{amssymb}
\usepackage{graphicx}
\usepackage{diagbox}
\restylefloat{figure}
\journal{Expert Systems with Applications}
\UseRawInputEncoding
\biboptions{authoryear}
\begin{document}
	\begin{frontmatter}

		\begin{titlepage}
			\begin{center}
				\vspace*{1cm}
				
				\textbf{ \large Algorithmic Trading Using Continuous Action Space Deep Reinforcement Learning}
				
				\vspace{1.5cm}
				
				Naseh Majidi$^{a}$ (naseh.majidi@sharif.edu), Mahdi Shamsi$^a$ (shamsi.mahdi@ee.sharif.edu), Farokh Marvasti$^a$ (fmarvasti@gmail.com) \\
				
				\hspace{10pt}
				
				\begin{flushleft}
					\small  
					$^a$ Faculty of Electrical Engineering, Sharif University of Technology, Azadi Ave, 1458889694 Tehran, Iran. \\

					\vspace{1cm}
					\textbf{Corresponding Author:} \\
					Farokh Marvasti \\
					Faculty of Electrical Engineering, Sharif University of Technology, Azadi Ave, 1458889694 Tehran, Iran. \\
					Tel: (98) 9123729799 \\
					Email: fmarvasti@gmail.com
					
				\end{flushleft}        
			\end{center}
		\end{titlepage}

\title{Algorithmic Trading Using Continuous Action Space Deep Reinforcement Learning}

\author[label1]{Naseh Majidi}
\ead{naseh.majidi@sharif.edu}

\author[label1]{Mahdi Shamsi}
\ead{shamsi.mahdi@ee.sharif.edu}

\author[label1]{Farokh Marvasti \corref{cor1}}
\ead{fmarvasti@gmail.com}

\cortext[cor1]{Corresponding author.}
\address[label1]{Faculty of Electrical Engineering, Sharif University of Technology, Azadi Ave, 1458889694 Tehran, Iran.}

\begin{abstract}
	Price movement prediction has always been one of the traders’ concerns in financial market trading. In order to increase their profit, they can analyze the historical data and predict the price movement. The large size of the data and complex relations between them lead us to use algorithmic trading and artificial intelligence. This paper aims to offer an approach using Twin-Delayed DDPG (TD3) and the daily close price in order to achieve a trading strategy in the stock and cryptocurrency markets. Unlike previous studies using a discrete action space reinforcement learning algorithm, the TD3 is continuous, offering both position and the number of trading shares. Both the stock (Amazon) and cryptocurrency (Bitcoin) markets are addressed in this research to evaluate the performance of the proposed algorithm. The achieved strategy using the TD3 is compared with some algorithms using technical analysis, reinforcement learning, stochastic, and deterministic strategies through two standard metrics, Return and Sharpe ratio. The results indicate that employing both position and the number of trading shares can improve the performance of a trading system based on the mentioned metrics.
\end{abstract}

\begin{keyword}
	Bitcoin \sep Algorithmic Trading \sep Stock Market Prediction \sep Deep Reinforcement Learning \sep Artificial Intelligence \sep Financial AI.
\end{keyword}

\end{frontmatter}
\section{Introduction}
\label{sec:intro}

Forecasting price movements in the financial market is a difficult task. According to the Efficient-Market hypothesis \citep{kirkpatrick2008analysis}, stock market prices follow a random walk process with unpredictable future fluctuations. When it comes to Bitcoin, its price fluctuates highly, which makes its forecasting challenging \citep{phaladisailoed2018machine}. Technical and fundamental analysis are two typical tools used by traders to build their trading strategies in the financial markets. According to price movement and trading volume, technical analysis provides trading signals \citep{murphy1999technical}. Fundamental analysis, unlike the former, examines related economic and financial factors to determine a security's underlying worth \citep{drakopoulou2016review}.

Humans and computers both perform data analysis. Although humans are able to keep an eye on financial charts (such as prices) and make decisions based on their past experiences, managing a vast volume of data is complicated due to various factors influencing the price movement. As a result, algorithmic trading has emerged to tackle this issue. Algorithmic trading is a type of trading where a computer that has been pre-programmed with a specific set of mathematical rules is employed \citep{theate2021}. There are two sorts of approaches in financial markets: price prediction and algorithmic trading. Price prediction aims to build a model that can precisely predict future prices, whereas algorithmic trading is not limited to the price prediction and attempts to participate in the financial market (e.g. choosing a position and the number of trading shares) to maximize profit \citep{hirchoua2021deep}. It is claimed that a more precise prediction does not necessarily result in a higher profit. In other words, a trader's overall loss due to incorrect actions may be greater than the gain due to correct ones \citep{li2019}. Therefore, algorithmic trading has been the focus of this study.

Classical Machine Learning (ML) and Deep Learning (DL), which are powerful tools for recognizing patterns, have been employed in various research fields. In recent years, using the ML as an intelligent agent has risen in popularity over the alternative of the traditional approaches in which a human being makes a decision. For two reasons, the ML and DL have enhanced the performance in algorithmic trading. Firstly, they can extract complex patterns from data that are difficult for humans to accomplish. Secondly, emotion does not affect their performance, which is a disadvantage for humans \citep{chakole2021}. However, there are two compelling reasons why the ML and DL in a supervised learning approach are unsuitable for algorithmic trading. Firstly, supervised learning is improper for learning problems with long-term and delayed rewards \citep{dang2019}, such as trading in financial markets, which is why Reinforcement Learning (RL), a subfield of ML, is required to solve a decision-making problem (trading) in an uncertain environment (financial market) using the Markov Decision Process (MDP). Secondly, in supervised learning, labeling is a critical issue affecting the performance of the final model. To illustrate, classification and regression approaches with defined labels may not be appropriate, leading to the selection of RL, which does not require labels and instead uses a goal (reward function) to determine its policy.

Recent studies have usually employed discrete action space RL to address algorithmic trading problems
\citep{chakole2021,jeong2019,shi2021,theate2021}, which compels traders to buy/sell a specific number of shares, which is not a useful approach in financial markets. On the contrary, the continuous action space RL is used in this study to let the trader buy/sell a dynamic number of shares. Furthermore, the results are compared to the TDQN algorithm \citep{theate2021}, two technical strategies, Buy/Sell and Hold algorithms, and some random and deterministic strategies in the presence of transaction costs. 

The main contributions of this work can be summarized as follows:

\begin{itemize}
	\item We have developed a novel continuous action space DRL algorithm (TD3) in algorithmic trading: this helps traders with managing their money while opening a position.
	\item This research aims both cryptocurrency and stock markets. 	 
\end{itemize}

The remainder of this paper is organized as follows: Section \ref{sec:backmater} provides a glossary of terms that readers will need to comprehend the rest of the paper. Section \ref{sec:related works} discusses financial market research that has been conducted using statistical methods, classical machine learning, deep learning, and reinforcement learning. The model elements are defined in Section \ref{sec:DRT}. Section \ref{sec:results} offers some baseline models and standard metrics, as well as the evaluation of the models. The final section discusses the findings and some recommendations for further study.

\section{Background Materials}\label{sec:backmater}
In this section, some terms are introduced in order to assist readers in understanding the rest of the article. A general definition of reinforcement learning and its elements are presented in the first part. The second one introduces model-free RL algorithms (Q-leaning and Deep Q-Network), while the last one provides a statistical test (T-test), which is required to compare results from two samples.

\subsection{Reinforcement Learning}\label{ssec:RL-back}
Reinforcement learning is one of the widely used machine learning approaches, which is composed of an agent, environment, reward, and policy. The RL agent interacts with its environment to learn a policy that maximizes the received reward. This procedure is quite similar to a situation in which someone is learning to trade in financial markets. The RL tries to solve a problem through an MDP that has four components: 
\begin{itemize}
	\item State Space: $ S $,
	\item Action Space: $ A $,
	\item Transition Probability between the states: $ P $,
	\item Immediate Reward: $ R (s, a) $.
\end{itemize}

\begin{figure}[!h]
	\centering
	\includegraphics[width=0.7\linewidth]{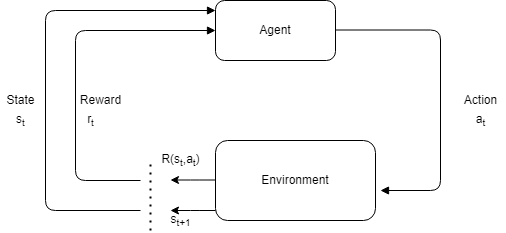}
	\caption{The RL process and components.}
	\label{fig:rl}
\end{figure}

The environment and what the agent observes at time $ t $ is represented by $ s_t $, which is used by the agent to take an action $ a_t $. The transition probability ($ P_{ss'}^a $) shows the probability of transitioning from the current state ($ s_t $) to the next one ($ s_{t+1} $) through an action, which is defined as:
$$P_{ss'}^a = P({s_{t + 1}} = s'|{s_t} = s,{a_t} = a).$$
When a transition occurs, the environment provides the agent an immediate reward $ R(s,a) $, indicating how much the taken action in the current state is beneficial or detrimental. The whole RL process is illustrated in Fig. \ref{fig:rl}.

There are two special features demarcating the RL from other types of learning (supervised, semi-supervised, and unsupervised), which are trial-and-error search and delayed rewards \citep{sutton2018}. To be more specific, an RL agent chooses different actions in the environment in order to find the optimal ones (trial-and-error search), and the future rewards also affect the current action (delayed rewards). Consequently, the purpose of the RL is to maximize the expectation of the discounted reward ($ G_t $) in order to obtain the optimal policy. The following mathematical formulation defines the mentioned reward:
\begin{equation}
	\label{eqn:G}
	G_t=\sum_{k=0}^{\infty}{\gamma^kr_{t+k+1}} = r_{t+1} + \sum_{k=1}^{\infty}{\gamma^kr_{t+k+1}}
\end{equation}
where $ r_{t+k+1} $ and $ \gamma $  denote the immediate reward at $ t+k+1 $ and discounted factor, respectively.

In order to reach this goal, the RL agent must take an action to reach a state that provides it with the highest average reward. Hence, a value function under the policy of $\pi$ ($ V_\pi $) is defined to represent this average reward \citep{sutton2018}:  
$${V_\pi }(s) = \mathbb{E}_{_\pi }\;\left\{ {{G_t}|{s_t}\; = \;s} \right\}\; \mathop  = \limits^{(1)}  \mathbb{E}{_\pi }\left\{ {{r_{t + 1}} + \;\sum\limits_{k = 1}^\infty  {{\gamma ^k}{r_{t + k + 1}}} {\rm{|}}{s_t}\; = \;s\;} \right\}\;\,\;{s_t} \in S,$$
$${V_\pi }(s) = \mathbb{E}{_\pi }\{ {r_{t + 1}} + \;\gamma {G_{t + 1}}|{s_t}\; = \;s\},$$
$${V_\pi }\left( s \right) = \sum\limits_a {\pi \left( {s|a} \right)} \sum\limits_{r,{s^\prime }} {P\left( {{s^\prime },r|s,a} \right)[r + \gamma \mathbb{E}{_\pi }\{ \;{G_{t + 1}}|{s_{t + 1}}\; = \;s^\prime \} \;],}$$
$$V_\pi\left(s\right)=\sum_{a}{\pi\left(s\middle| a\right)\sum_{r,s^\prime}{P\left(s^\prime,r\middle| s,a\right)\left[r+\gamma V_\pi\left(s^\prime\right)\right]}},$$
where $\pi(s|a) $ is the policy function showing the probability of choosing the action $ a $ in the state $ s $. The term action-value function ($ Q_\pi(s,a) $) is also used in the RL, representing the average received reward in the state $ s $ when the action $ a $ is taken under the policy of $\pi$:
\begin{equation}
	\label{eqn:q}
	{Q_\pi }(s,a) =  \mathbb{E}{_\pi }\{ {G_t}|{s_t}\; = \;s,\;{a_t} = a\} \; = \mathbb{E}{_\pi }\{ {r_{t + 1}} + \;\gamma {G_{t + 1}}|{s_t}\; = \;s\;,\;{a_t} = a\} .
\end{equation}

By applying Eq. \ref{eqn:q}, the RL agent aims to achieve the optimal policy. To be more specific, a policy maximizing the value of $ Q $ or $ V $ is optimal and is shown by $ \pi^* $. Furthermore, $ Q^* $ and $ V^* $ denote the optimal values of $ Q $ and $ V $, respectively:
$${Q^ * }\left( {s,a} \right) = \mathop {\max }\limits_\pi  {Q_\pi }\left( {s,a} \right),$$
\begin{equation}
	\label{bell_q}
	{Q^ * }(s,a) =\sum\limits_{s',r} {P(s',r|s,a)[r + \gamma \mathop {max}\limits_{a'} {Q^*}(s',a')]} ,
\end{equation}
$${V^ * }\left( s \right) = \mathop {max}\limits_\pi  {V_\pi }\left( s \right)\; = \;\mathop {max}\limits_a {Q_{\pi^*} }\left( {s,a} \right),$$
\begin{equation}
	\label{bell_v}
	{V^ * }\left( s \right) = \mathop {max}\limits_a \sum\limits_{s',r} {P(s',r|s,a)[r + \gamma {V^ * }(s^\prime)]} ,
\end{equation}
where Eq. \ref{bell_q} and \ref{bell_v} are known as Bellman optimality equations \citep{sutton2018} for $ V $ and $ Q $, owning the transition probability in their formulas.

\subsection{Q-Learning and Deep Q-Network (DQN)}\label{ssec:Qlearning}
Q-learning is an off-policy and model-free RL technique that updates its Q-values using Temporal Difference (shown in Eq. \ref{eqn:qlearning}). The significant advantage of using this approach is that the Q-values can be obtained without the need for explicit knowledge of the transition probabilities:
\begin{equation}
	\label{eqn:qlearning}
	Q\left(s_t,a_t\right)=Q\left(s_t,a_t\right)+\alpha\left[r_t+\gamma\mathop {max}\limits_a {Q\left(s_{t+1},a\right)}-Q\left(s_t,a_t\right)\right],
\end{equation}
where $ \alpha $ is the learning rate of the algorithm. There is a lookup table (Q-Table) in the Q-learning approach whose rows and columns are associated with states and actions, respectively. The agent takes an action in the environment, and the Q-table is updated according to the Eq. \ref{eqn:qlearning}. Moreover, there is a trade-off between exploration and exploitation to allow the agent to take a wide range of actions in order to achieve the optimal policy. To put it in another way, the $ \epsilon $-greedy strategy is adopted, in which an action is performed based on the Q-table with a probability of $ 1-\epsilon $ (exploitation), or that is taken randomly with a probability of $ \epsilon $ (exploration), where $ 0\leq\epsilon\leq1 $.

Mostly, when Q-learning is utilized to solve real-world problems, the number of states grows too large; hence, Q-learning may not be the best solution. To overcome such a problem, DQN \citep{mnih2015}, which is a combination of Q-leaning and Deep Neural Network, is used to estimate the Q-values ($Q_{\theta}^{pred}$ ) from the input (state). In other words, a target Q-value ($Q_{\theta^-}^{target}$) is calculated using Eq. \ref{eqn:qtarget}, and the network tries to predict the Q-values converging to the target Q-values  as defined in Eq. \ref{eqn:qtarget}. In order to have more stable training, the target network, which calculates target Q-values, is isolated from the network (main network), which calculates the Q-values. The weights of the target network are constant, and the main network's weights are copied into the target network after $ M $ iterations. As a result, back-propagation does not occur in the target network. Additionally, when the agent takes an action in the environment, a tuple containing $ (s,a,r,s^\prime) $ is saved in a buffer (experience reply buffer). Finally, some samples from the buffer are chosen to update the main network’s weights using Eq. \ref{eqn:lossf} and \ref{eqn:backpro}:

\begin{equation}
	\label{eqn:qtarget}
	Q_{{\theta ^ - }}^{target} = \;\left\{ \begin{array}{l}
		{r_{t + 1}} + \gamma \mathop {\max }\limits_{{a^\prime }} {Q_{{\theta ^ - }}}\left( {{s_{t + 1}},{a^\prime }} \right)\,\,\,\,\,\,t < T\\
		{r_t}\,\,\,\,\,\,\,\,\,\,\,\,\,\,\,\,\,\,\,\,\,\,\,\,\,\,\,\,\,\,\,\,\,\,\,\,\,\,\,\,\,\,\,\,\,\,\,\,\,\,\,\,\,\,\,\, t = T
	\end{array} \right.,
\end{equation}
\begin{equation}
	\label{eqn:lossf}
	L\left(\theta\right)=\frac{1}{N}\sum_{i=1}^{N}{\left(Q_{\theta^-}^{target}-Q_\theta^{pred}\right)^2,}
\end{equation}
\begin{equation}
	\label{eqn:backpro}
	\theta\gets\theta-\alpha\nabla L\left(\theta\right).
\end{equation}

\subsection{T-test}\label{ssec:t-test}
A statistical test is necessary to make a conclusion regarding the attributes of a population (mean and variance) using a limited number of observations. Such a test generalizes the result derived from the observations with a confidence factor ${\gamma _{conf}} = 1 - {\alpha _{conf}}$, where there are two hypotheses.

T-test compares the mean of two dependent populations as one of the described tests. The hypotheses of the one-sided version of the T-test are as follows:
$$\left\{ \begin{array}{l}
	{H_0}:\,\mu {}_x \ge \mu {}_y\\
	{H_1}:\,\mu {}_x < \mu {}_y
\end{array} \right.,$$
where $ \mu_x $ and $ \mu_y $ are the means of the populations. In order to carry out the test using samples $ X_i $ and $ Y_i $, $ D_i $ must be derived using:
$$ D_i=X_i-Y_i. $$

Then, statistics ($ T_0 $) and P-value are obtained using the following equations, and the $ H_0 $ is rejected if $ \alpha _{conf} $ is less than P-value:
$$ T_0=\frac{\bar{D}}{\sqrt{\frac{S_D^2}{n}}}, \,\,\,\,	P-value\ =\ P(T>T_0),$$
where $ \bar{D}  $, $ S_D^2 $ are respectively the mean and variance of $ D_i $, $ n $ is the size of the samples, and $ T $ is the variable of T-distribution with the degree of freedom $ n-1 $.

\section{Related Works}\label{sec:related works}
This section presents the latest studies on trading in financial markets utilizing statistical and machine learning techniques (Classical ML, DL, and RL). The RL review is set apart from previous techniques since the RL is the primary focus of this research.

\subsection{Statistical, Classical Machine Learning, and Deep Learning}\label{ssec:S&ML}

Recent research trend in financial market prediction has focused on the statistical learning, ML, and DL methods. For stock index prediction, the Autoregressive Integrated Moving Average (ARIMA) method has been utilized, which obtained acceptable results for short-term forecasting \citep{ariyo2014}. ARIMA has been compared to Long Short-Term Memory (LSTM) and Recurrent Neural Network (RNN), and it has been shown that the LSTM surpasses both the RNN and ARIMA \citep{mcnally2018}. Furthermore, while endogenous and exogenous variables have been utilized as the input to a neural network, the gated-recurrent network with recurrent dropout achieved a lower Root Mean Squared Error (RMSE) than the LSTM network \citep{dutta2020}. Although the Convolutional Neural Networks (CNNs) are often used for object detection and image recognition \citep{pathak2018,traore2018deep}, a combination of wavelets and the CNNs outperformed the LSTM, CNN and Multi-Layer Perceptron (MLP) for forecasting S\&P500 trend \citep{persio2016}. In \citep{hoseinzade2019}, the 3D-CNN and the 2D-CNN have been compared to technical indicators strategies, and in most cases, the CNN-based algorithms outperformed in terms of the Sharpe ratio and the CEQ return. The CNN has been surpassed by a combination of the LSTM and CNN since they could detect both dependencies and local patterns in price time series \citep{alonso2020}. Another architecture that has been used to forecast the Bitcoin prices is autoencoder. The Stacked Denoising Auto Encoder outperformed the Support Vector Machine, Back Propagation Neural Network, and Principal Component Analysis-based Support Vector Regression in terms of Mean Absolute Percentage Error, Root Mean Squared Error, and Directional Accuracy  \citep{liu2021}. In terms of network depth, deeper networks have been shown to perform better than shallow ones \citep{liu2022}.

\subsection{Reinforcement Learning}\label{ssec:RL-Related}
Algorithmic trading and financial market prediction issues are also addressed by the RL approaches. In order to embed the OHLCV (open, high, low, close price, and volume) data into the states of RL problems, Authors in \citep{chakole2021} employed two approaches. The first approach utilized the k-means method to divide states into $ n $ groups, while the second one quantized the percentage change between close and open prices into six levels. In most situations, the first approach, employing clustering, outperformed the second one, according to the results. 

On the other hand, Deep RL is a popular solution to the problems that researchers have lately adopted. Raw data can be fed into a neural network, estimating Q-values or actions (buy, hold, and sell). A state-of-the-art DRL algorithm called DQN has been employed to handle financial trading decisions. In order to predict the trading share, a Deep Neural Network (DNN) was included in that system \citep{jeong2019}. Researchers in \citep{yang2020deep} also used a system with more than three actions. They used the outputs of the Advantage Actor-Critic (A2C), Proximal Policy Optimization (PPO), Deep Deterministic Policy Gradient (DDPG), and an ensemble of them to allow traders to choose several levels as an action. The ensemble model outperforms all the individual models in terms of the Sharpe ratio. However, the PPO model obtained the highest Return among the aforementioned models. Researchers in \citep{li2019} have also looked at dynamic trading assets, where a model with seven actions outperformed a model with only three. 

When comparing DRL-based algorithms to ML-based ones, a DRL agent with a core of Double Deep Q-Network beats LSTM and SVM models \citep{shi2021} in terms of Return. Furthermore, the agent outperforms Buy and Hold. 

Since DNN is at the core of the DRL agent, the techniques used in DNN models to avoid overfitting are taken into account. Batch Normalization, Dropout, and Gradient Clipping were employed by \citep{theate2021} to help with its optimization. 

Authors in \citep{betancourt2021} have created a portfolio management technique that can trade on a dynamic number of assets as a version of algorithmic trading. This has been shown to be beneficial when it comes to the introduction of new coins into the Bitcoin market.

\section{Methodology}\label{sec:DRT}
In this section, a new approach of trading in financial markets is considered in which a continuous action space DRL addresses the trading problem. Generally, each DRL problem consists of four components that must be defined first: State, Environment, Action, and Reward. Hence, as follows, we present our proposed approach by properly defining those components. Finally, the TD3, which is addressed to solve the DRL problem, is briefly explained.

\subsection{State Space}\label{ssec:State}
Candlestick datasets, which include open, close, high, low price, and volume, are provided in several time steps, such as weekly, daily, and hourly. Traders extract a trading signal from the data (short-term and long-term) according to their aims. The most common one among traders is the daily resolution. 

\begin{figure}[!h]
	\centering
	\includegraphics[width=0.7\linewidth]{"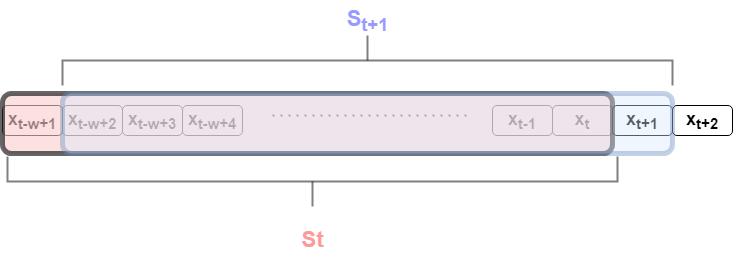"}
	\caption{The sliding window creating the states. }
	\label{fig:state}
\end{figure}

As the price data are not stationary, the percentage change is applied to the close price in order to obtain stationary data:
$$ x_t=100 \times \frac{p_t-p_{t-1}}{p_{t-1}} , $$								
where $ p_t $ and $ x_t $ are the close price and its percentage change at time $ t $, respectively. As shown in Fig. \ref{fig:state}, a sliding window is used to define the state space, so that at time $ t $, the current state (red) is made up of the last $ w $ data samples ($ x_{t-i}; \;i\in\{0,1,...,w-1\} $), where $ w $ is the length of the sliding window. The next state (blue) is formed when the window moves by one step.

\subsection{Environment and Action Space}\label{ssec:Env,Action}
Traders make decisions based on the data provided in financial markets. Long, Hold, and Short are considered to be the three actions (indicated by $ a_t $) that traders are allowed to take at time $ t $. When traders open a "Long" position, it implies they anticipate the asset's value to rise, while a "Short" one suggests the trader expects the asset's value to decrease. It is obvious that the profit is realized when the forecast is correct. The term "Hold" refers to the trader remaining in the position and making no changes to the asset's shares. 

An agent starts with initial cash ($ c_0 $) and trades under a specified trading strategy in a financial market, where the agent opens a position at the beginning of the day and closes it at the end of that. Since the agent's action space is continuous, it picks an action in the interval $\left[-1,1\right]$, where $ -1 $ and $ +1 $ represent that the total amount of cash is paid to open a short and long position, respectively. In addition, if $ a_t\in(-1,1) $, the agent opens a position through holding $h_t = \left|a_t\right|c_{t-1}$ amount of the cash, and the remainder of that is unchanged. Obviously, positive and negative actions reveal long and short positions, respectively. Then the number of held shares is calculated by $n_t=\frac{h_t}{p_t}$, and after closing the position, the cash is calculated through
$$c_t=c_{t-1}-h_t+max\Big(n_t\;(p_{t+1} \pm p_t)+h_t-\frac{n_t\times TC \times p_t}{100},0\Big),$$
where $ p_t $ is the close price at time $ t $, and $ TC $ is the transaction cost of each action. Furthermore, $ "+" $ denotes a long position, whereas $ "–" $ denotes a short one.

\subsection{Reward}
When an agent interacts with the environment, it is guided by its received feedback (reward function), which helps it achieve the optimal policy. Clearly, a positive reward encourages the agent to execute the action, but a negative one discourages the agent from taking the action. In general, the Return function $({Return}_t=\frac{c_t\ -\ c_{t-1}}{c_{t-1}})$  is one of the most popular rewards for the objective function of an RL approach. The logarithm of the Return function is employed here, allowing us to control the final Return of the entire period ($ R_T $), as illustrated in Eq. \ref{eqn:logret}:
$$ 	r_t=log\left(1+{Return}_t\right)=log\left(\frac{c_t}{c_{t-1}}\right), $$
\begin{equation}
	\label{eqn:logret}
	R_T=\sum_{i=1}^{T}{r_i=}\sum_{i=1}^{T}{log(\frac{c_i}{c_{i-1}})=log(\prod_{i=1}^{T}{\frac{c_i}{c_{i-1}})\ =\ log(\frac{c_T}{c_0})}},
\end{equation}
where $ r_t $ and $ T $ are the immediate reward and the length of the whole trading period, respectively.

\subsection{Twin-Delayed Deep Deterministic (TD3)}\label{ssec:TD3}
Unlike the DQN method described in Subsection \ref{ssec:Qlearning}, which is a discrete action space algorithm, the TD3 \citep{fujimoto2018} is a continuous action space approach that is suitable for our situation. It is an actor-critic technique in which the actor network estimates the policy based on the states via a neural network, and the critic ones use both the states and the action, chosen by the actor network, in order to estimate the value function. The TD3 includes two critic networks and two critic target networks, unlike the DDPG \citep{silver2014}, to overcome the problem of overestimation \citep{fujimoto2018}. Nevertheless, the TD3’s actor network is the same as the DDPG’s, with one actor and one actor target network.

Turning to the details, several episodes with random actions are first run to explore the environment, and the transitions $  (s,a,r,s’) $ are stored in a reply buffer. Following this, in each episode according to current state ($ s $), an action ($\pi_\varphi\left(s\right)$) is chosen based on actor network and is added to the exploration noise ($\varepsilon_1$):
$$a=\pi_\varphi\left(s\right)+\varepsilon_1,\ \ \ \ \  \varepsilon_1 \sim \mathcal{N}\left(0,\sigma\right),$$
where $a$ and $\sigma$ are the noisy action and the standard deviation of the noise, respectively. Since the agent has less familiarity with the environment at the beginning of learning, this exploration noise should has a significant amount to help the agent explore the environment. Furthermore, this noise should diminish exponentially as the number of episodes grows since the agent becomes more acquainted with the environment than it was before:
$$\sigma=\sigma_{end}+\left(\sigma_{ini}-\sigma_{end}\right)exp\left(-\frac{N_{episode}}{D_{\sigma}}\right),$$
where $\sigma_{ini}$ and $ \sigma_{end} $ are respectively the initial and final values of the $ \sigma $, $
N_{episode} $ is the number of episodes, and $ D_{\sigma} $ is the decay parameter of the exponential function.

Then, the transitions $ (s,a,r,s^\prime) $ are again stored in the replay buffer. According to a small batch of the them, the next actions ($ a^\prime = \pi_{\varphi^\prime}(s^\prime) $) are resulted from the next states ($ s^\prime $) via the actor target network. Meanwhile, a Gaussian noise ($ \varepsilon_2 $), as the policy noise, is added to the next action. Then, both the next action and policy noise are limited between two defined values using the Clip function which clamps the first argument between the second and third one:
$$\widetilde{a}={Clip(\pi}_{\varphi^\prime}(s^\prime)+\varepsilon_2\ ,\ a_1\ ,\ a_2),$$
$$\varepsilon_2 \sim Clip\left(\mathcal{N}\left(0,\widetilde{\sigma}\right),-K,K\right),$$ 
where $ \widetilde{a} $ is the final value of the target action, $ a_1 $ and $ a_2 $ are the minimum and maximum possible values of the target action, $ \widetilde{\sigma} $ is the standard deviation of policy noise, and $ -K $ and $ K $ are the minimum and maximum possible values of the noise. According to the above argument for exploration noise, an exponential decay approach is employed for both the standard deviation of the policy noise ($ \widetilde{\sigma} $) and noise limiter ($ K $):
$$\widetilde{\sigma}={\widetilde{\sigma}}_{end}+\left({\widetilde{\sigma}}_{ini}-{\widetilde{\sigma}}_{end}\right) exp\left(-\frac{N_{episode}}{D_{\widetilde{\sigma}}}\right),$$
$$K=K_{end}+\left(K_{ini}-K_{end}\right) exp\left(-\frac{N_{episode}}{D_K}\right).$$

Afterward, $ y $ is computed using the equation as follows, which is in turn used in the optimization of critic models:
$$y\ =\ r\ +\ \gamma\ \mathop {\min }\limits_{i = 1,2} {Q_{{\theta ^\prime }_i}}\left( {{s^\prime },a} \right),$$
where $ y $ is the ultimate target Q-value, $ Q_{{\theta ^\prime }_i}( {s^\prime },a) $ is the Q-value of the $ i^{th} $ critic target model, $ r $ is the immediate reward, and $ \gamma $ is the discount factor. The next step involves updating the weights of critic models done by Eq. \ref{eqn:backpro}, where the critic loss is computed by
$$ L\left( {{\theta }} \right) = {N^{ - 1}}\sum\limits_{j = 1}^2 {\sum\limits_{i = 1}^N ( \,{y_i} - {Q_{i,{\theta _j}}}\left( {{s^\prime },a} \right)} ), $$
where $ N $ is the batch number. Then, after every $ N_0 $ iterations, the weights of actor network are updated accordingly:
\begin{equation}
	\label{eqn:gradient of l}
	\nabla_\varphi J\left(\varphi\right)=N^{-1}\sum_{i=1}^{N}{\nabla_aQ_{\theta1}(s,a)|\ _{a=\pi_\varphi\left(s\right)}}\nabla_\varphi\pi_\varphi\left(s\right),
\end{equation}
$$ \varphi\gets\varphi+\alpha\nabla_\varphi J\left(\varphi\right). $$

The gradient clipping technique is applied to Eq. \ref{eqn:gradient of l} in order to stabilize the training stage. Finally, the target networks are updated every $ N_0 $ iterations using 
$${{\theta}^\prime}_i\gets\tau\theta_i+\left(1-\tau\right){\theta^\prime}_i,$$
$${\varphi^\prime}_i\gets\tau\varphi_i+\left(1-\tau\right){\varphi^\prime}_i,$$
where $ 0\leq\tau\leq1 $.

\section{Experimental Results}\label{sec:results}
The simulations, provided in this section, have been done on Python 3.7 and the neural network models have been built by PyTorch 1.12.1. The dataset was acquired from the free version of the Yahoo Finance API in Python, containing the close price of Bitcoin (BTC) from 2014-10-15 to 2020-01-01 and Amazon (AMZN) from 2010-01-01 to 2021-06-01 with the daily time step. The experiment has been divided into three stages: training, validation, and testing, in which we divided the dataset into 80\%, 10\%, and 10\% portions, respectively.

First, nine models and two metrics are briefly introduced in order to be used in the performance evaluation of the TD3. It is then evaluated in the stock market using the AMZN. In this scenario, the TD3 is compared to two discrete algorithms in order to demonstrate the superiority of our proposed continuous algorithm. ّFinally, this algorithm is examined in the cryptocurrency market (Bitcoin). In all cases, the agent starts with 100,000\$ as its initial cash.

\subsection{Baseline Models and Metrics}\label{ssec:model and metric}
There are two random models (Random-C, Random-D), four deterministic models (Buy and Hold, Sell and Hold, Long, and Short), two technical-indicator-based models (MRMA\footnote{Mean Reversion Moving Average} and TFMA\footnote{Trend Following Moving Average}), and one DRL-oriented model (TDQN) \citep{theate2021}, which are explained in Table \ref{table:models}. Moreover, the models are evaluated using two standard metrics (Return and Sharpe ratio) described in Table \ref{table:met}.

\begin{table}[!h]
	\centering
	\caption{Baseline models descriptions.}
	\label{table:models}
	\begin{tabular}{lcp{8cm}}
		\hline
		Model & Approach & Description  \\
		\hline
		{Random-C (RC)} & Random & Continuous Uniform Distribution of $ [-1,1] $\\
		
		{Random-D (RD)} & Random & Discrete Uniform Distribution of $ \left\{ -1,1\right\} $ \\
		
		{Buy and Hold (BH)} & Deterministic & Opens a long position and holds it till the end of trading period \\
		\
		{Sell and Hold (SH)} & Deterministic & Opens a short position and holds it till the end of trading period \\
		
		{Long} & Deterministic & Opens a long position and closes it after a 1-day interval \\
		
		{Short}  & Deterministic & Opens a short position and closes it after a 1-day interval \\
		
		{MRMA} & Technical Indicator & Assumes that the price goes back to the average of the price \\
		
		{TFMA} & Technical Indicator & Follows the previous trend of the price \\
		
		{TDQN} & DRL & DQN\\
		\hline
	\end{tabular}
\end{table}
\begin{table}[!h]
	\centering
	\caption{Performance metrics.}
	\label{table:met}
	\begin{tabular}{lc}
		\hline
		Metric & Formula \\
		\hline
		{Return (\%)} & $ 100\times \frac{\textnormal{Final cash - Initial cash}}{\textnormal{Initial cash}} $\\
		
		{Sharpe ratio} & $\sqrt{\textnormal{Number of trading days in a year}} \times \frac{\textnormal{Average of Returns}}{\textnormal{Standard deviation of Returns}}$\\
		\hline
	\end{tabular}
\end{table}
\subsection{Amazon}\label{ssec:AMZN}
\begin{figure}
	\centering
	\includegraphics[width=0.7\linewidth]{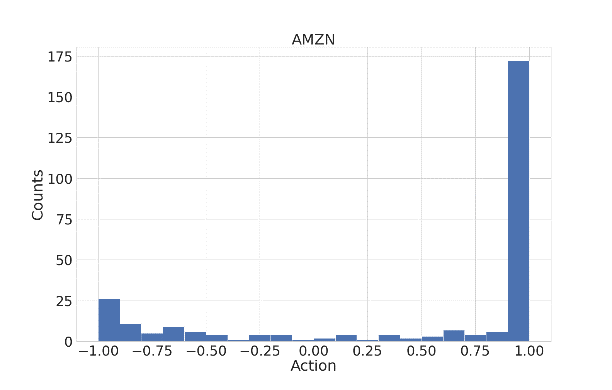}
	\caption{The histogram of Amazon market actions in the TD3 algorithm.}
	\label{fig:amznhist}
\end{figure}

\begin{figure}
	\centering
	\includegraphics[width=0.7\linewidth]{"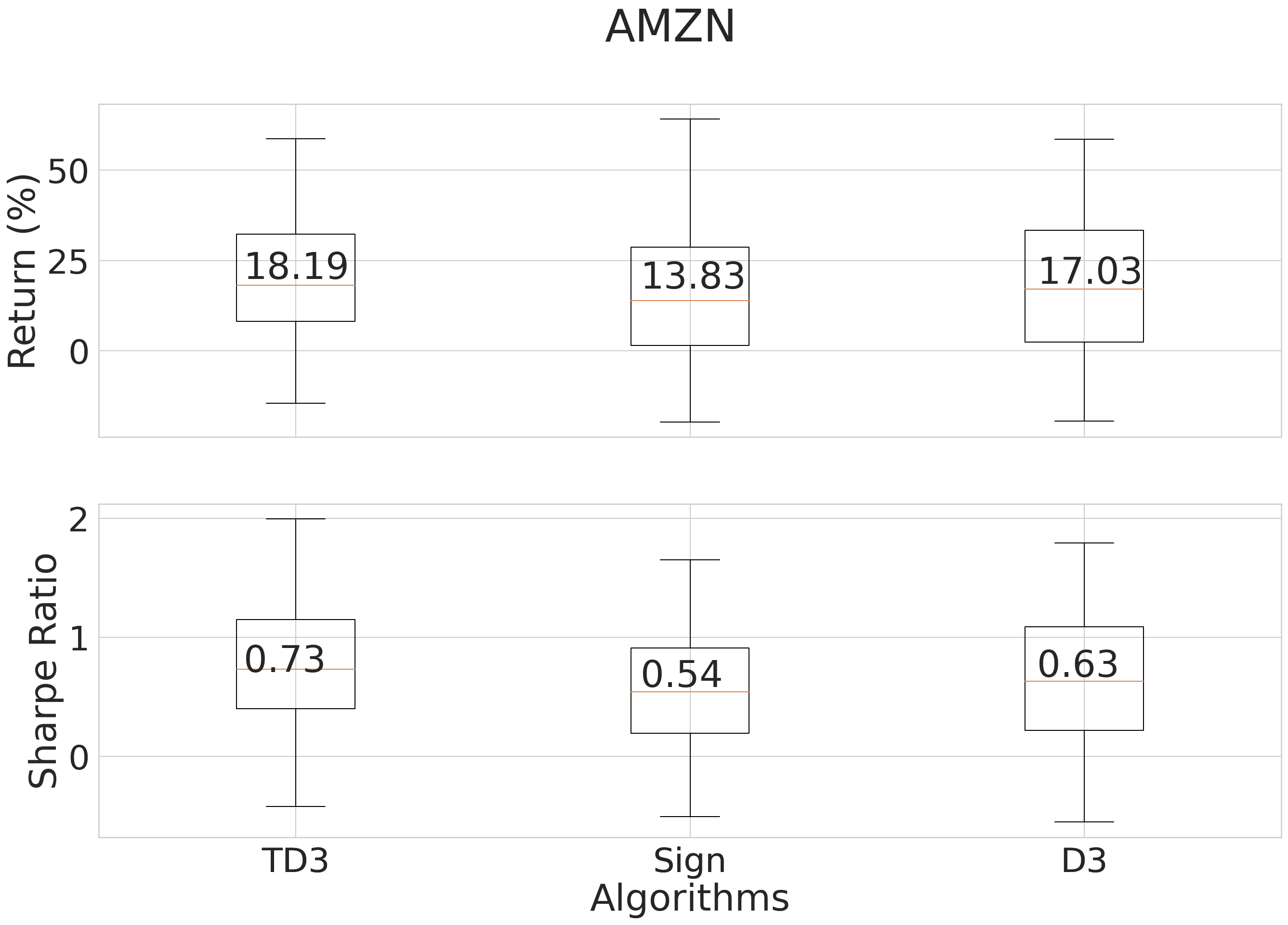"}
	\caption{Box-plot compares the TD3, Sign, and D3. }
	\label{fig:amznaction-1}
\end{figure}

\begin{figure}
	\centering
	\includegraphics[width=0.7\linewidth]{"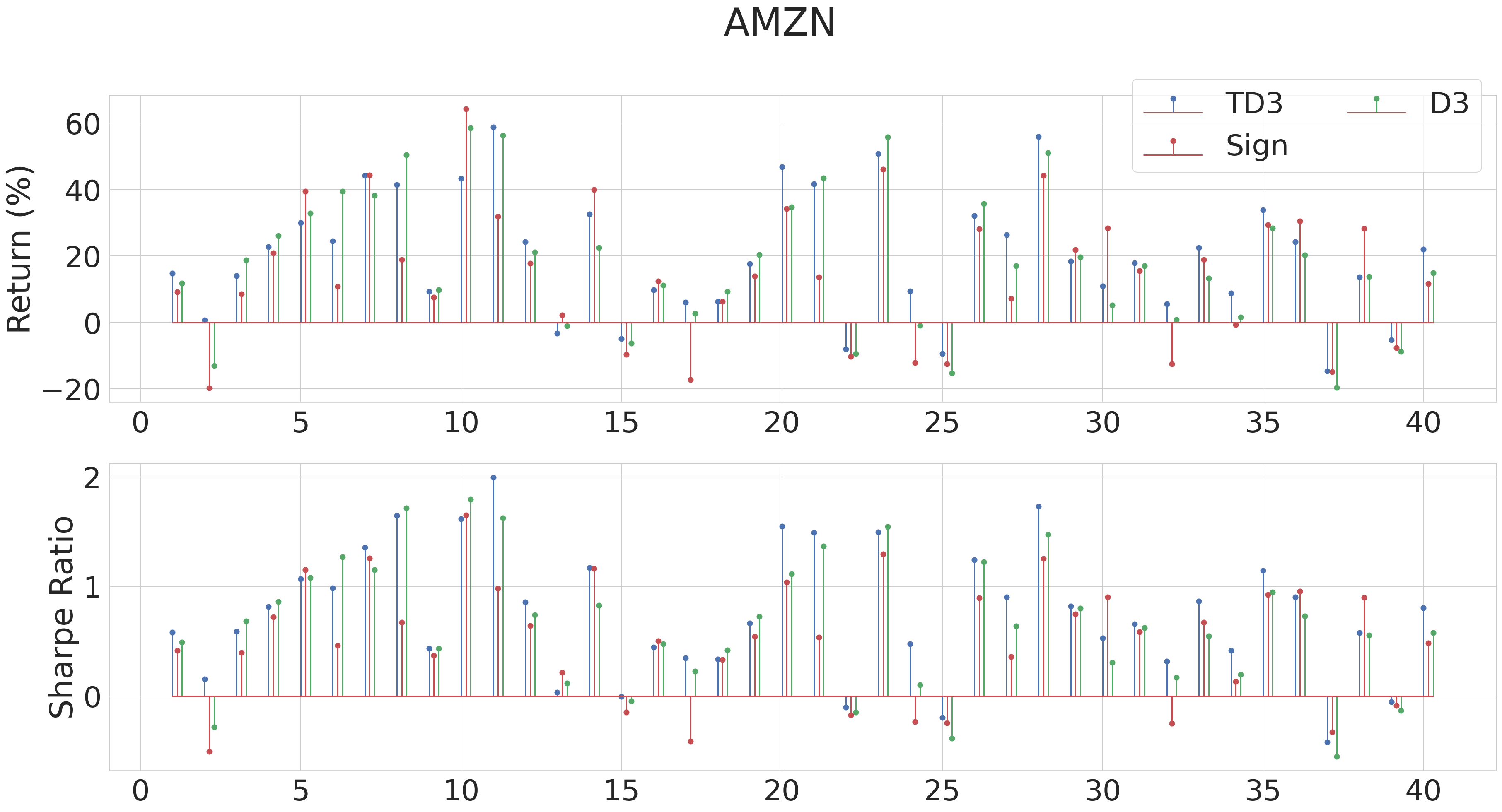"}
	\caption{All 40 results of the TD3, Sign, and D3. }
	\label{fig:amzncomp-1}
\end{figure}

This part evaluates the performance of the TD3 in the AMZN market. Fig. \ref{fig:amznhist} shows the histogram of the TD3's actions in AMZN trading. According to the figure, the values of the actions are between +1 and -1, allowing us to claim that the action distribution is continuous. 
In order to make a fair comparison between continuous (TD3) and discrete action space approaches, Sign (Eq. \ref{eqn:sign}) and D3 (Eq. \ref{eqn:d3}), two discrete algorithms, are defined as follows:

\begin{equation}
	\label{eqn:sign}
	{f_{Sign}}(a) = \left\{ \begin{array}{l}
		- 1\,\,\,\,\,\,\,\,\,\,\,\,\,\,\,\,\,\,\,\,\,\,\,a <  =  0\\
		+ 1\,\,\,\,\,\,\,\,\,\,\,\,\,\,\,\,\,\,\,\,\,\,\,a > 0
	\end{array} \right.
\end{equation}

\begin{equation}
	\label{eqn:d3}
	{f_{D3}}(a) = \left\{ \begin{array}{l}
		- 1\,\,\,\,\,\,\,\,\,\,\,\,\,\,\,\,\,\,\,\,\,\,\,a <  =  - \frac{1}{3}\\
		0\,\,\,\,\,\,\,\,\,\,\,\,\,\,\,\,\,\,\,\,\,\,\,\,\,\, - \frac{1}{3} < a <  = \frac{1}{3}\\
		+ 1\,\,\,\,\,\,\,\,\,\,\,\,\,\,\,\,\,\,\,\,\,\,\,a > \frac{1}{3}
	\end{array} \right.
\end{equation}
where $ a $, $ {f_{Sign}}(a) $, and $ {f_{D3}}(a) $ are the actions of the TD3, Sign, and D3 algorithms, respectively. These aforementioned algorithms are simulated 40 times, and Fig. \ref{fig:amznaction-1} indicates the results. The figure illustrates the superiority of a continuous action space algorithm over the discrete ones by showing that the medians of Return and Sharpe ratio in the TD3 algorithm is higher than those in both discrete algorithms. Further comparing these two algorithms reveals that although the Sign algorithm may achieve a higher Return, this increase was accompanied by a decline in the Sharpe ratio value. In other words, the TD3 has a greater chance of realizing a better Sharpe ratio than the other two algorithms. The possibility of reducing the amount of invested money in various situations can be the cause of that (for instance, risky circumstances). The simulation results of the algorithms are also separately depicted in Fig. \ref{fig:amzncomp-1}. According to the figure, the Return of the TD3 was greater than Sign in 75\% of the cases and the D3 in 60\% . These values ​​for the Sharpe ratio were 80\% and 70\%, respectively.


For Return and Sharpe ratio, the results of the mean comparison test ($ \alpha _{conf} = 0.01 $) are provided in Table \ref{table:t_test}. $ X_i $ and $ Y_i $ are the results of the discrete algorithms (Sign or D3) and the TD3, respectively. We deduce that the null hypothesis is rejected in all cases with a certainty of 99\%, showing that the TD3 performs better than both Sign and the D3 because the P-value in each case is less than $ \alpha _{conf} $, except for the Return test between the D3 and TD3. However, it can be rejected with above 94\% certainty.

\begin{table}[!h]
	\centering
	\caption{T-test results.}
	\label{table:t_test}
	\begin{tabular}{lcccc}
		\hline 
		\multicolumn{1}{l}{Algorithms} & \multicolumn{2}{c}{TD3 and Sign}
		& \multicolumn{2}{c}{TD3 and D3}
		\\
		\cline{2-5}
		{}& $ T_{0} $  & P-value & $ T_{0} $  & P-value \\
		\hline
		Return & -2.84 & 0.003 & -1.66 & 0.052  \\ 
		
		Sharpe ratio & -4.41 & 0.00004 & -3.95 & 0.00016 \\
		\hline
		
	\end{tabular}
\end{table}
Table \ref{table:amzn_com} compares the performance of the algorithms in AMZN market. The BH outperforms the other algorithms in terms of Return and Sharpe ratio. Moreover, the TD3 and TDQN are the second and third best algorithms, which implies that the DRL is a useful approach in such a case, and a continuous model performs better than a discrete one.

\begin{table}[!h]
	\centering
	\caption{Algorithm comparison in AMZN market.}
	\label{table:amzn_com}
	\begin{tabular}{lcc}
		\hline
		Algorithm & Return (\%)  & Sharpe ratio \\
		\hline
		{TD3} & 9.3 & 0.43 \\
		
		{TDQN} & 7.03 & 0.36 \\
		
		{BH} & \textbf{35.4} & \textbf{1.05} \\
		
		{SH} & -35.3 & -0.48 \\
		
		{MRMA} & -20.0 & -0.42 \\
		
		{TFMA} & -10.9 & -0.20 \\
		
		{Long} & 7.4 & 0.37 \\
		
		{Short} & -31.6 & -2.02 \\
		
		{Random-C} & -27.2 & -1.50 \\
		
		{Random-D} & -29.1 & -0.87 \\
		\hline		
	\end{tabular}
\end{table}

\subsection{Bitcoin}\label{ssec:btc}
\begin{figure}[!h]
	\centering
	\includegraphics[width=0.7\linewidth]{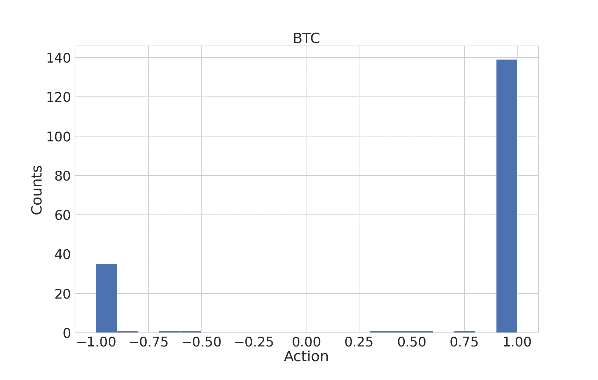}
	\caption{The histogram of Bitcoin market actions in the TD3 algorithm.}
	\label{fig:btchist}
\end{figure}

The histogram of the TD3 actions in the BTC scenario is illustrated in Fig. \ref{fig:btchist}. Almost all the actions are either +1 or -1, demonstrating that the distribution of the actions is discrete; thus, there is no comparison between continuous and discretized models in this part. The TD3 algorithm is compared to the baseline ones in Table \ref{table:btc_com}. The first point highlighted by the table is that the TD3 and TDQN outperform other models, indicating that the DRL can also be a applicable approach in this market. Furthermore, the TD3 surpasses TDQN, demonstrating the benefits of an actor-critic model to an only-critic one.

\begin{table}[!h]
	\centering
	\caption{Algorithm comparison in BTC market.}
	\label{table:btc_com}
	\begin{tabular}{lcc}
		\hline
		Algorithm & Return (\%)  & Sharpe ratio \\ 
		\hline
		{TD3} & \textbf{57.5} &\textbf{1.53}\\
		
		{TDQN} & 29.4 & 1.39 \\
		
		{BH} & -28.4 & -1.06 \\
		
		{SH} & 28.2 & 1.37 \\
		
		{MRMA} & 1.4 & 0.29 \\
		
		{TFMA} & -13.2 & -0.34 \\
		
		{Long} & -43.7 & -1.4 \\
		
		{Short} & -0.7 & 0.31 \\
		
		{Random-C} & -2.3 & 0.02 \\
		
		{Random-D} & -0.1 & 0.33 \\
		\hline		
	\end{tabular}
\end{table}

\section{Conclusion and Future Works}\label{sec:conclusion}
In recent years, the use of machine learning in algorithmic trading has become increasingly widespread, which is why this research aimed to present an experiment tackling this issue. Since both the number of trading shares and position are crucial in trading, the continuous action space DRL was addressed in exchange for the discrete one. According to the experiments on AMZN data, the chosen approach (TD3) outperformed discrete action space approaches (TDQN, Sign, and the D3) in terms of Return and Sharpe ratio. Moreover, the superiority of the DRL-oriented approaches over the algorithms which are common among the traders (in most cases) demonstrated the capacity of DRL in this field in both AMZN and BTC markets.

However, there are some certain limitations to this method, which should be addressed in future research. To begin with, modifying the reward function can improve the agent's performance. This modification should make the agent's purpose more similar to a real trader's. In other words, there should be a barrier in order to prevent massive losses in trading activities. To manage this issue, choosing the Sharpe ratio as the reward function can be a potential solution. Second, since traders use a combination of methods to create their strategy, the TD3's performance can be improved by using an ensemble of methods from diverse fields, such as machine learning and technical-indicator-based methods. Finally, combining data from many sources may be advantageous. These adjustments were introduced to make the method closer to how real traders create a profitable strategy.

\bibliography{myrefs}

\end{document}